\documentclass[10pt, a4paper,x11names]{article}
\usepackage{setspace}
\usepackage[]{misc/lrec-coling2024}
\usepackage{times}
\usepackage{xcolor}
\usepackage{graphicx}
\usepackage[T1]{fontenc}
\usepackage[utf8]{inputenc}
\usepackage{microtype}
\usepackage{booktabs}
\usepackage{todonotes}
\usepackage{amsmath,amssymb}
\usepackage{cleveref}
\usepackage{xspace}
\usepackage{listings,algorithm, algorithmicx, algpseudocode}
\usepackage{soul} %
\usepackage{caption}
\usepackage{multirow,multicol}
\usepackage{float}
\usepackage{array}
\usepackage{bm}
\usepackage{xfrac}
\usepackage[shortlabels]{enumitem}

\usepackage[normalem]{ulem} %

\usepackage{tabularx}

\crefname{lstlisting}{listing}{listings}
\Crefname{lstlisting}{Listing}{Listings}
\crefname{equ}{equation}{equations}
\Crefname{equ}{Equation}{Equations}
\Crefname{algorithm}{Algorithm}{Algorithms}
\crefname{example}{example}{examples}
\Crefname{example}{Example}{Examples}
\crefname{prompt}{prompt}{prompts}
\Crefname{prompt}{Prompt}{Prompts}
\DeclareCaptionType{example}[Example][List of examples]
\DeclareCaptionType{prompt}[Prompt][List of prompts]
\DeclareCaptionType{equ}[Equation][List of equations]

\usepackage{courier}
\usepackage{marginnote}

\definecolor{TodoColor}{rgb}{1,0.7,0.6}

\newcommand\comet[2][]{COMET$_{#2}^\textrm{#1}$\xspace}
\newcommand{\hrefEmail}[2]{\href{mailto:#1}{\color{black}{#2}}}

\newcommand{\offsetminus}{\hspace{-1.2mm}-}
\newcommand{\plusA}{\ensuremath{\bm{\bm{+}}}}
\newcommand{\minusA}{\ensuremath{\bm{\bm{-}}}}
\newcommand{\btau}{\ensuremath{\bm{\tau}}\xspace}
\newcommand{\observation}{\raisebox{-0.4mm}{\includegraphics[height=3.1mm,trim=0 2mm 0 2mm]{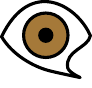}}\xspace}
\newcommand{\PE}{\ensuremath{^{\textrm{PE}}}\xspace}
\newcommand{\PEx}{\ensuremath{^{\textrm{PE}}}}

\newcommand{\blackindicator}[1]{\scalebox{0.7}[#1]{$\blacksquare$}}

\newcommand{\fcost}{\ensuremath{\textsc{Cost}}}
\newcommand{\futility}{\ensuremath{\textsc{Util.}}}
\newcommand{\fadd}{\ensuremath{\textsc{Add}}}
\newcommand{\fmove}{\ensuremath{\textsc{Promote}}}

\definecolor{FindingsColor}{gray}{0.85}

\let\svthefootnote\thefootnote
\newcommand\blankfootnote[1]{%
  \let\thefootnote\relax\footnotetext{#1}%
  \let\thefootnote\svthefootnote%
}

\title{Quality and Quantity of Machine Translation\\ References for Automatic Metrics}

\name{
\makebox[4cm]{Vilém Zouhar}
\quad 
\makebox[4cm]{Ondřej Bojar}
}

\address{
    \makebox[4cm]{ETH Zürich}
    \quad
    \makebox[4cm]{Charles University} \\
    \makebox[4cm]{\texttt{\hrefEmail{vzouhar@ethz.ch}{vzouhar@ethz.ch}}}
    \quad
    \makebox[4cm]{\texttt{\hrefEmail{bojar@ufal.cuni.cz}{bojar@ufal.cuni.cz}}}
}

\abstract{
Automatic machine translation metrics typically rely on \textit{human} translations to determine the quality of \textit{system} translations.
Common wisdom in the field dictates that the human references should be of very high quality.
However, there are no cost-benefit analyses that could be used to guide practitioners who plan to collect references for machine translation evaluation.
We find that higher-quality references lead to better metric correlations with humans at the segment-level.
Having up to 7 references per segment and taking their average (or maximum) helps all metrics.
Interestingly, the references from vendors of different qualities can be mixed together and improve metric success.
Higher quality references, however, cost more to create and we frame this as an optimization problem: given a specific budget, what references should be collected to maximize metric success.
These findings can be used by evaluators of shared tasks when references need to be created under a certain budget.
}

\begin{document}
\maketitleabstract

\section{Introduction}
\label{sec:introduction}

\footnotetext[0]{\href{https://github.com/ufal/optimal-reference-translations}{\fontsize{9.2}{10}\selectfont github.com/ufal/optimal-reference-translations} 
\href{https://huggingface.co/datasets/zouharvi/optimal-reference-translations}{\fontsize{9.2}{10}\selectfont hf.co/datasets/zouharvi/optimal-reference-translations} 
}

Machine translation systems are robustly evaluated through human annotation.
This is non-scaleable and non-replicable \citep{10.1162/tacl_a_00437} for settings such as shared tasks where a number of teams submit automatic translations of the same testset.
Automatic metrics aim to provide a cheap and replicable solution.
Given the translation and possibly the source and reference segments, they produce a score that correlates with what a human annotator would predict.
There is support and evidence for not using references \citep{lommel2016blues} in metrics, i.e. quality estimation \citep{specia2018machine,rei-etal-2021-references}.
Still, most of the commonly used metrics (\Cref{sec:setup}) require human reference translations \citep{freitag-etal-2023-results}.
These metrics work by comparing either the overlap on the surface-level \citep[e.g. BLEU,][]{papineni-etal-2002-bleu}, of semantic representations \citep[e.g. COMET,][]{rei-etal-2020-comet} or some downstream task \citep[e.g. MTEQA,][]{krubinski-etal-2021-just}.

Humans also do not always arrive at perfect translations and thus the quality of the references themselves varies \citep{castilho2018approaches}.
In cases of very poor translations, such as non-translation,\footnote{Text left untouched in the source language.} the reference-based metrics would clearly fail.
While low-quality references are known to decrease the metric correlations \citep{freitag-etal-2023-results}, the extent of this effect and interactions with other phenomena remains unclear.
Many automatic machine translation metrics support multiple references for a single translation natively or by using an aggreation such as the average.
For phrase-based MT and BLEU, the trade-off between the number of references vs. the test set size was studied by
\citet[][Section 5]{bojar-EtAl:2013:WMT}, concluding that a single-reference test set of 3000 sentences can be comparable to 6--7 references with just 100--200 test sentences.
The usefulness of multiple references was later disputed \citep{freitag-etal-2020-bleu} for state-of-the-art system evaluation and some recent metrics do not even support multiple references.
Additionally, a professional experienced translator is likely to produce a better translation than an average crowd-worker.
However, the cost of a high-quality human translation is likely also much higher.

In this paper, we aim to quantify the trade-off between reference \textbf{quality, quantity} and \textbf{cost} for segment-level automatic metric performance.
We base our experiments on a small-scale English$\rightarrow$Czech dataset with multiple references of varying qualities.

We pose research questions with immediate implications for practitioners.
The short answers here are only summaries.
\vspace{-3mm}
\begin{itemize}[left=4.2mm,label={},noitemsep,topsep=0mm]
\item
\item[\textbf{Q:}] Are higher-quality references useful for automatic evaluation?
\item[\textbf{A:}] Low-quality should be avoided. Too much investment has diminishing returns.  \hfill (Sec.~\ref{sec:ref_quality})
\\[-4mm]

\item[\textbf{Q:}]
Are multiple references useful? 
\item[\textbf{A:}] Yes. Averaging or taking the maximum across reference improves the metrics. \hfill (Sec.~\ref{sec:multiple_ref})
\\[-4mm]

\item[\textbf{Q:}]
How to allocate the budget? 
\item[\textbf{A:}] By not focusing exclusively on either quality or quantity of references but their combination.
This can be computed by \Cref{alg:ref_allocation}, given a list of vendors and their attributes. \hfill (Sec.~\ref{sec:budget})
\end{itemize}

\section{Related Work}
\label{sec:related_work}

Reference quality is known to affect machine translation evaluation.
\citet{freitag-etal-2023-results} note that very low-quality references reduce metric success.
This stands in contrast to the pre-neural machine translation era where the reference quality did not play an important role in certain settings \citep{hamon-mostefa-2008-impact}.
This is likely caused by the much higher quality of systems being compared.
\citet{vernikos-etal-2022-embarrassingly} hypothesize that ambiguous and vague references are the culprits of metric success deterioration.
Additionally, \citet{freitag-etal-2020-bleu} study how to avoid low-quality references in human translation campaigns.

The BLEU metric \citep{papineni-etal-2002-bleu} was intended to be used with multiple metrics, which was only rarely put in practice over the years.
Nevertheless, newer and more sophisticated methods exist to incorporate them \citep{qin-specia-2015-truly}.
Our results from \Cref{fig:ref_quantity} confirm the older observations of \citet{finch2004does} or \citet[][Section 5]{bojar-EtAl:2013:WMT} who study the effect of the reference count on metric performance.
Finally, multiple references can be used in training better machine translation systems \citep{madnani-etal-2008-multiple,zheng-etal-2018-multi,khayrallah-etal-2020-simulated,mi2020improving} or for analyzing model uncertainty \citep{pmlr-v80-ott18a} or evaluation uncertainty \citep{zhang-vogel-2004-measuring,zhang2010significance,fomicheva-etal-2020-multi}.
It is also used outside of machine translation for measuring consensus \citep{vedantam2015cider}.

The budget allocation algorithm is reminiscent of active learning or data selection.
In machine translation, this is limited to selecting training examples \citep{haffari-etal-2009-active,gonzalez-rubio-etal-2012-active,van-der-wees-etal-2017-dynamic,shi-huang-2020-robustness,10.1162/coli_a_00473}.
We focus on algorithmic data selection for higher-quality \textit{evaluation}.
We aim to complete similar works on practical advice on machine translation evaluation.
\citet{kocmi-etal-2021-ship,kocmi2024navigating} study the reliability of metrics from the perspective of deployment decisions.
We show that the configuration of references can make these metrics stronger or weaker on segment-level.

\section{Setup}
\label{sec:setup}
\vspace{-2mm}

To evaluate the effect of references on automatic machine translation evaluation, we need data with controlled references and reference-based metrics.

\vspace{-3mm}
\paragraph{Optimal Reference Translations.}
\citet{zouhar2023evaluating} re-annotate a subset of the English$\rightarrow$Czech testset from the News domain of the WMT2020 campaign \citep{barrault-EtAl:2020:WMT1}.  
New references were created by translating the original source in 4 different human settings ranging from generic translation vendors to translatology academics following a novel protocol leading to so-called ``optimal reference translations'' \citep{ort-SaS:2023}.
This phase was followed by a human annotation and post-editing phase performed by 11 annotators of varying professionalities.

\citet{zouhar2023evaluating} study whether the human quality of the references is really the highest achievable one.
They stop short of evaluating the impact of this on machine translation evaluation.
We re-purpose their data and system submissions from \citet{barrault-EtAl:2020:WMT1}.
We refer to the references, from lowest to highest quality of the source, as R1, R2, R3, and R4.
Specifically, R1 to R2 come from standard translation vendors,\footnote{Nevertheless, based on observations of \citet{kloudova2021detecting}, R1 are to a large extent post-edits of one of the participating systems.} R3 is high-quality translation vendor, and R4 is the work of translatologists (the optimal reference).
See \Cref{tab:dataset_overview} for basic statistics.

\footnotetext[2]{
\newcommand{\smalltt}[1]{\texttt{\fontsize{8.8}{9}\selectfont #1}}
\smalltt{BLEU|\#:1|c:mixed|e:yes|tok:13a|s:exp} \\
\smalltt{chrF|\#:1|c:mixed|e:yes|nc:6|nw:0|s:no} \\
\smalltt{TER|\#:1|c:lc|t:tercom|nr:no|pn:yes|as:no}
}

\begin{table}[htbp]
\centering
\resizebox{\linewidth}{!}{
\begin{tabular}{ll}
\toprule
Source segments \& documents & $160$ \,\&\, $20$ \\
Average source segment length & $34$ tokens \\
Reference segments & $160{\times} 4=640$ \\
Reference post-editing & $160{\times}4{\times} 11 = 7040$ \\
Systems \& system segments & $13$ \,\,\&\,\, $160{\times}13 = 2080$ \\
\bottomrule
\end{tabular}
}
\caption{Overview of the used dataset.}
\label{tab:dataset_overview}
\vspace{-5mm}
\end{table}

\paragraph{Automated Metrics.}\hspace{-4mm}\footnotemark\quad
For the metrics, we use BLEU \citep{papineni-etal-2002-bleu}, chrF \citep{popovic-2015-chrf}, TER \citep{snover-etal-2006-study}, \href{http://huggingface.co/unbabel/wmt20-comet-da}{\comet{20}} \citep{rei-etal-2020-comet}, its referenceless version \href{https://huggingface.co/Unbabel/wmt20-comet-qe-da}{\comet[QE]{20}}, and its updated iteration \href{http://huggingface.co/unbabel/wmt22-comet-da}{\comet{22}} \citep{rei-etal-2022-comet}, and \href{https://huggingface.co/lucadiliello/BLEURT-20-D12}{BLEURT} \citep{sellam-etal-2020-bleurt}.
We select these as a representative set of widely-used string-matching and trainable metrics.

\paragraph{Metric Evaluation.}
We focus on and evaluate metric success at the segment-level (``sentence''-level) by correlating the metric scores with human scores using Kendall's \btau.\footnote{
$\btau = (\textrm{\#concordant} - \textrm{\#discordant})/\textrm{\#pairs}$; read more on the definition in \citet{machacek:wmt14metrics:2014}.
}
Each translation receives a human score and automatic metric scores which are correlated.
This is the standard segment-level evaluation adopted by the WMT Metrics Shared Task \citep{freitag-etal-2021-results,freitag-etal-2022-results,freitag-etal-2023-results}.
In our case (WMT2020), the human segment-level judgments were created from Direct Assessment \citep{graham:da:2016} judgements following the ``\textsc{da}RR'' conversion as described by \citet{metrics:2020}: Candidate translations from MT systems were scored on their own, independently of other candidates. For each pair of judgements of candidates translating the same source, we construct one golden-truth item of pairwise comparison if the two individual scores differ by more than 25\% absolute.
As \citet{metrics:2020}, we believe that this difference in the judgement is big enough to trust the simulated pairwise comparison.

\begin{table}[t]
\centering
\resizebox{\linewidth}{!}{
\begin{tabular}{lcccc}
\toprule
\bf Metric & \bf R1 & \bf R2 & \bf R3 & \bf R4 \\
\midrule
BLEU         & 0.082 \blackindicator{0.09} & 0.103 \blackindicator{0.25} & 0.109 \blackindicator{0.30} & 0.103 \blackindicator{0.25} \\
chrF         & 0.090 \blackindicator{0.16} & 0.125 \blackindicator{0.42} & 0.128 \blackindicator{0.44} & 0.123 \blackindicator{0.41} \\
TER          & 0.082 \blackindicator{0.09} & 0.092 \blackindicator{0.17} & 0.114 \blackindicator{0.34} & 0.105 \blackindicator{0.27} \\
\comet{20} & 0.172 \blackindicator{0.79} & 0.176 \blackindicator{0.82} & 0.185 \blackindicator{0.89} & 0.181 \blackindicator{0.85} \\
\comet{22} & 0.189 \blackindicator{0.92} & 0.195 \blackindicator{0.96} & 0.191 \blackindicator{0.93} & 0.192 \blackindicator{0.94} \\
BLEURT       & 0.159 \blackindicator{0.69} & 0.156 \blackindicator{0.66} & 0.199 \blackindicator{0.99} & 0.178 \blackindicator{0.83} \\
\cmidrule{1-1}
\bf Average & 0.129 \blackindicator{0.46} & 0.141 \blackindicator{0.55} & 0.154 \blackindicator{0.65} & 0.147 \blackindicator{0.59} \\
\cmidrule{1-1}
\comet[QE]{20} & \multicolumn{4}{c}{0.171 \blackindicator{0.77}}\\
\bottomrule
\end{tabular}
}
\caption{Segment-level Kendall's \btau between automatic metrics and human scores. The metrics are computed with respect to each of the four references.
The black boxes indicate the value visually and comparable across both columns and rows.
\observation~The R3 translation yields the best results as the reference, despite not being the optimal translation from the human perspective.}
\label{tab:ref_quality_orig}
\vspace{-5mm}
\end{table}

\begin{table}[t]
\vspace{1em} %
\centering
\resizebox{\linewidth}{!}{
\begin{tabular}{lcccc}
\toprule
\bf Proficiency & \bf R1\PE-R1 & \bf R2\PE-R2 & \bf R3\PE-R3 & \bf R4\PE-R4 \\
\midrule
\bf Layman & \plusA 0.019 & \plusA 0.011 & \plusA 0.011 & \plusA 0.011 \\
\bf Student & \plusA 0.009 & \plusA 0.005 & \plusA 0.001 & \minusA 0.002 \\
\bf Professional & \plusA 0.025 & \plusA 0.011 & \plusA 0.004 & \plusA 0.002 \\
\bottomrule
\end{tabular}
}
\caption{Difference in Kendall's $\tau$ between using original translations (in \Cref{tab:ref_quality_orig}) and their post-edited versions. The post-editing comes from translators on different levels. The correlations are averaged across all metrics; see \Cref{tab:ref_quality_pe_full,tab:ref_quality_pe_fullplus} for per-metric breakdowns. 
\observation~In most cases, using post-edited versions improves metric performance.}
\label{tab:ref_quality_pe}
\end{table}

\section{Experiments}

\subsection{Reference Quality is Important}
\label{sec:ref_quality}

As stated in \Cref{sec:setup}, we have access to four human translations of varying quality.
In \Cref{tab:ref_quality_orig}, we show the metric success measured by correlation with human scores.
The metrics stay the same but the references they use are changed.
Across both string-matching and parametrized model-based metrics, R1, the worst human translation, leads to the worst metric performance.
The best performance is achieved with R3, a standard professional translation.
Notably, it is not R4 which was created by professional translatologists and was also the most expensive one.
This can be explained by the presence of translation shifts, which occur more frequently on this professionality level, but can negatively impact the utility of the reference \citep{fomicheva2017role}.
Translation shifts in general refer to deviation from the original structure or meaning.
For our new references, the translatologists paid attention to preserve the meaning but they often restructured the sentences.
They did this to avoid translationese as much as possible and to express the subtleties of information structure (given vs. new information) which is natively expressed via Czech word order.
These boosted word order differences make it harder for automatic metrics to match the, rather translationese, candidate and the reference.
We anticipate that more fluent large language model-based MT could sound less translationese and the ``optimal reference translations'' will serve better in this setting.
See \Cref{sec:qualitative_analysis} for an example and analysis.

A simple way of improving a translation is to post-edit (refine) it, which is cheaper that translating it from scratch \citep{daems2020post,zouhar-etal-2021-neural}.
Moreover, \citet[][Figure 7]{bojar-EtAl:2013:WMT} show that such post-edited references lead to a better performance of BLEU, because ``every n-gram mismatch indicates an error''.
With standard references, an n-gram mismatch often means just lack of reference coverage.
However, such post-edited references need to be ideally created for each evaluated MT system.
In our case, the post-edits were created starting from \emph{human} reference translations Rx.
We mark them Rx\PE and use them as references for the automatic metrics in \Cref{tab:ref_quality_pe}.
The post-editors are either laymen with knowledge of both languages, students of translatology, or professional translators.
While the proficiency level plays a role, in most cases the post-edited translations serve as better references.
\Cref{tab:ref_quality_pe_full} below lists the raw metric score changes in a closer detail.

\begin{table}[t] %
\centering
\resizebox{0.95\linewidth}{!}{
\begin{tabular}{lcccc}
\toprule
\bf Aggregation & \bf R3 & \bf R\{3,4\} & \bf Rx & \bf Rx\PE \\
\midrule
\bf Average & 0.154 \blackindicator{0.60} & 0.159 \blackindicator{0.64} & 0.166 \blackindicator{0.68} & 0.164 \blackindicator{0.67} \\
\bf Max & 0.154 \blackindicator{0.60} & 0.155 \blackindicator{0.61} & 0.165 \blackindicator{0.68} & 0.167 \blackindicator{0.70} \\
\bottomrule
\end{tabular}
}
\caption{Average performance of metrics with multiple references. See \Cref{tab:ref_quantity_full} for per-metric breakdown. \observation~All aggregation methods improve the performance over the best single one, R3.}
\label{tab:ref_quantity}
\end{table}

\subsection{Multiple References are Useful}
\label{sec:multiple_ref}

The previous section provided an analysis of how individual references affect metric performance.
In many situations, however, multiple references are available.
While some metrics, such as BLEU, support multiple references natively, one can also aggregate them using either segment-level averages or maxima (i.e. compute multiple scores for each segment and take the average or maximum).
In \Cref{tab:ref_quantity} we consider three setups: (1) two high-quality references, R3 and R4, (2) all human translations, Rx, or (3) all post-edited human translations, Rx\PE.
Across all metrics, this segment-level aggregation improves the correlation with humans, especially in the case of using the original four human translations.
Taking the maximum and not the average has the advantage that there exists a specific reference which yields that particular score.
The maximum also reflects the spirit of automated evaluation: measuring some similarity between the candidate and reference translations.
With more references, taking the maximum corresponds to first finding the most similar reference.
We include all subsets as references in \Cref{tab:ref_quantity_full}.

\begin{figure}[t]
\centering
\includegraphics[width=\linewidth]{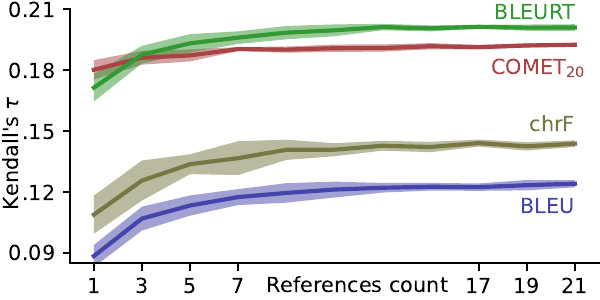}
\caption{Metric performance with multiple sampled references from the pool of the original human translations and their post-edited versions.
Confidence t-test intervals indicate 99\% confidence of the mean (of 10 samples) being in the shaded area. \observation~Biggest advantage is gained from at least three references and taking their segment-level average (max aggregation not shown).}
\label{fig:ref_quantity}
\end{figure}

To systematically study the effect of reference count on metric performance, in \Cref{fig:ref_quantity} we randomly sample $x$ references from the whole pool of original and post-edited translations, irrespective of their quality.
The biggest gains in metric performance are achieved until seven references and further gains are negligible, which is in line with the observations of \citet[][Section 5]{bojar-EtAl:2013:WMT}.

\begin{table}[htbp]
\centering
\resizebox{\linewidth}{!}{
\begin{tabular}{lcccccc}
\toprule
\bf Metric & \bf R1 & \bf R2 & \bf R4 & \bf R3 & \bf R1\PE & \bf R3\PE\\
\midrule
BLEU         & 24.2 & 31.5 & 27.3 & 37.1 & 23.9 & 31.0 \\
chrF         & 55.7 & 60.3 & 56.1 & 63.0 & 54.5 & 58.4 \\
TER          & \offsetminus63.3 & \offsetminus53.0 & \offsetminus59.4 & \offsetminus48.7 & \offsetminus64.1 & \offsetminus58.9 \\
COMET$^{20}$ & 65.5 & 68.9 & 61.0 & 68.2 & 60.4 & 61.4 \\
COMET$^{22}$ & 84.6 & 84.9 & 83.6 & 84.8 & 83.6 & 83.7 \\
BLEURT       & 61.3 & 66.1 & 64.5 & 68.8 & 61.6 & 64.9 \\
\bottomrule
\end{tabular}
}
\caption{Raw average scores across metrics and references. TER scores are flipped to make higher numbers be better.
The columns are sorted by quality of references from worst to best as reported in \Cref{tab:ref_quality_orig}.
\observation For most metrics, higher absolute metric scores correspond to better evaluation (numbers are growing from left to right), except for post-edited human references Rx\PE which serve better as references (are more to the right) but lead to lower absolute metric scores.}
\label{tab:ref_scores}
\vspace{-3mm}
\end{table}

\subsection{References and Metric Scores}
\label{sec:ref_scores}

To understand the effect of different metrics, we show the average \textit{raw} scores of each metric in \Cref{tab:ref_scores}.
While it appears that the higher the raw score, the better the metric performance (low score of R1 and high scores of R3 and R4), this trend does not explain the improvements of using the post-edited versions, e.g. as R1\PE over R1, or R3\PE over R3.
In fact, the post-edited versions always lead to lower raw scores.
This could be the result of either further translation shifts as the post-edits are based on a translation and not the source or additional (fully justified) corrections in the references which lead to fewer matches with the candidates.

\subsection{Allocating a Budget for References}
\label{sec:budget}

Usually, it is simple to gather many source sentences and let multiple systems translate them.
Evaluating all of them using human annotators is unattainable but running automatic metrics is not.
However, these require references, which are also costly.
It remains unclear how many references and of which quality to obtain to achieve the most reliable automatic quality assessment under a given budget.

\begin{figure}[htbp]
\centering
\includegraphics[width=\linewidth]{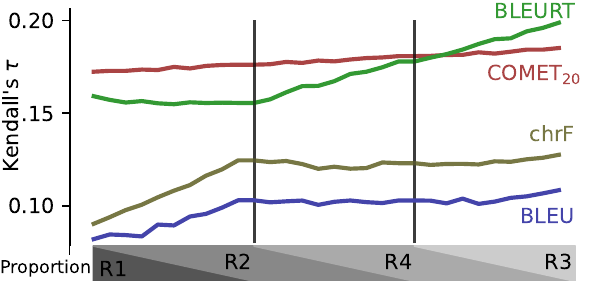}
\caption{Metric performance with references (ordered by usefulness) from mixed sources (e.g. 25\% R1 and 75\% R2; rightmost is 100\% R3). \observation~Mixing references does not hurt any metric.}
\label{fig:ref_mix}
\vspace{-4mm}
\end{figure}

\begin{algorithm*}
\caption{
Budget Allocation for References\\
\textbf{Input:} Source segments $S$, levels $L$, cost function $\fcost: L \rightarrow \mathbb{R}^+$, utility function\\
\null\hspace{10.5mm} 
$\futility: L \rightarrow \mathbb{R}^+$, tradeoff hyperparameter $\lambda \in [0, 1]$, temperature $t > 0$, budget $B \in \mathbb{R}^+$.\\
\textbf{Output:} Assignment $R : L \rightarrow 2^S$. \\
\textbf{Note:} \Cref{fig:budget_allocation} contains a patience mechanism instead of exit on error.
}
\label{alg:ref_allocation}
\setstretch{1.0}
\begin{algorithmic}[1]
\State $L \gets \Call{Sort}{L, \fcost}$
\State $R[L_0] \gets S$; \qquad $O \gets R$ \Comment{Assign everything to the cheapest level at first.}
\While{$\sum_{l\in L} |R[l]|\cdot \Call{\fcost}{l} < B$ \, $\wedge$ \, \textbf{no exception}}
    \State $O \gets R$
    \State $a \sim \Call{Sample}{\fmove: \lambda, \fadd: 1-\lambda}$ \Comment{Select action.}
    \State $X^+ \gets \{\langle s, l\rangle | l\in L, s\in S\setminus R[l] \}$ \Comment{Samples that could be added to $R[l]$.}
    \State $X^- \gets \{\langle s, l\rangle | l\in L, s\in R[l] \}$ \Comment{Samples that could be removed from $R[l]$.}
    \newline
    \If{$a = \fadd$}
        \State $x, l \sim \Call{Sample}{\{\langle x, l\rangle: \frac{\sigma(\futility(l)-\fcost(l))^{\sfrac{1}{t}}}{Z} | x, l \in X^+ \}}$
        \Comment{Sample a segment to add.}
        \State $R[l] \gets R[l] \cup \{x\}$ \Comment{Commit transaction.}
        \vspace{2mm}
    \ElsIf{$a = \fmove$}
        \vspace{2mm}
        \State $x^+, l^+ \sim \Call{Sample}{\{\langle x, l\rangle: \frac{\sigma(\futility(l)-\fcost(l))^{\sfrac{1}{t}}}{Z} | x, l \in X^+ \}}$ \Comment{Sample a segment to add.}
        \newline
        \null\hfill\Comment{Sample where to move from.}
        \State $\,\_\,\,\,, l^- \sim \Call{Samp.}{\{\langle x, l\rangle: \frac{\sigma(\fcost(l)-\futility(l))^{\sfrac{1}{t}}}{Z} | x, l \in X^-, x{=}x^+ \wedge \futility(l^-){<}\futility(l^+) \}}$
        \State $R[l^+] \gets R[l^+] \cup \{x^+\}$; \qquad $R[l^-] \gets R[l^-] \setminus \{x^+\}$ \Comment{Commit transaction.}
    \EndIf
\EndWhile; \qquad \Return $O$
\end{algorithmic}
\end{algorithm*}

\begin{figure*}[htbp]
\centering
\vspace{-1mm}
\includegraphics[width=0.9\linewidth]{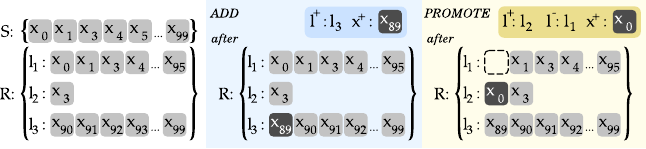}
\vspace{-2mm}
\caption{Illustration of two operations from \Cref{alg:ref_allocation}. The initial state is on the left. Then, a new segment x$_{89}$ \colorbox{DodgerBlue3!30}{is added} to the $l_3$ level. Lastly, the segment x$_{0}$ \colorbox{Goldenrod2!50}{is promoted} from $l_1$ to $l_2$.}
\label{fig:ref_allocation_illustration}
\vspace{-2mm}
\end{figure*}

\paragraph{Can references be mixed?}
To assess what types of configurations of references \textit{can} lead to the most reliable automatic evaluation, we first validate if references can be meaningfully mixed.
For example, if it is viable that 75\% of the sources can have references from a cheaper vendor R1 and 25\% from a higher-quality but more expensive vendor R3.
This is different from \Cref{tab:ref_quantity} where each segment had exactly two references from the same two sources.
In \Cref{sec:ref_scores}, we show that using lower-quality references R1 leads to lower absolute metric scores (e.g. BLEU = 24.2) as opposed to higher-quality ones R3 (e.g. BLEU = 37.1).
This holds across all metrics.
\citet{bojar-kos-marecek:2010:Short} observe that lower BLEU scores are less reliable, but they refer to the range of BLEU $<$ 20.
It is thus questionable if BLEUs at 20--40 correlate differently with human MT quality judgements.
In \Cref{fig:ref_mix}, we mix some of the references together for evaluation, but staying at single-reference evaluation.
Despite the varying absolute scores of metrics under different references, as explored in \Cref{sec:ref_scores}, mixing of multiple references leads to an almost perfectly linear combination of the endpoint metric performances.
The biggest gains in this respect are obtained by BLEURT, chrF and BLEU, while \comet{20} is almost unaffected.
There is no formal guarantee that the mix of score distributions will not lower the overall Kendall's $\tau$.
Nevertheless, a positive conclusion is that if there is budget to only translate 25\% of segments with high quality, it should be done and it can only improve the overall evaluation reliability.

\paragraph{Budget Allocation Algorithm.}

We provide a heuristic stochastic algorithm to find an assignment of source segments $S$ to be translated by vendors of different costs and qualities within a specific budget.
For the current dataset, we set the cost of a segment in R1, R2, R3, and R4 to be 1, 1, 2, and 3, respectively.
Their quality (or ``fitness'' for the purpose of automatic evaluation) were set to 1, 2, 4, and 3 based on our observations in \Cref{sec:ref_quality}.
\Cref{alg:ref_allocation} contains a hyperparameter $\lambda$ that controls whether the budget will be allocated more towards having multiple references per-segment or more towards having fewer but higher-quality references per-segment, and the temperature $t$ than controls the the sampling randomness.

We formalize the problem with a segment cost $\fcost(l)$ for a reference on level $l{\in}L$ and the utility $\futility(l)$.
The levels might correspond to translation vendors which have costs and qualities.
In our case, $\fcost = \{\textrm{R1:}\,1, \textrm{R2:}\,1, \textrm{R3:}\,2, \textrm{R4:}\,3 \}$ and $\futility = \{\textrm{R1:}\,1, \textrm{R2:}\,2, \textrm{R3:}\,4, \textrm{R4:}\,3 \}$.
Given a set of source sentences $S$, the goal is to assign the segments to different levels \textrm{R1} \dots \textrm{R4}.
The same segment can be assigned to different quality levels at once, leading to multiple references for that segment.
The selection should maximize performance of a particular metric on a number of systems but needs to fit under a fixed budget $B$, i.e. $\sum_l |R_l|\cdot \fcost(l) \leq B$.
In our setup, to preserve fair comparison, each segment needs to have at least one reference.
This is because the smaller the testset, the easier it is to achieve higher but spurious correlations.
Therefore, $\bigcup_{l\in L} R_l = S$.
The formalization explicitly allows for parts of the testset to be translated multiple times but requires the budget to cover at least the full test set with the cheapest references.
This requirement can be fulfilled by subsampling the testset, as commonly done in WMT evaluation campaigns \citep[][inter alia]{kocmi-etal-2023-findings}.

The pseudocode is provided in \Cref{alg:ref_allocation} and explanatory illustration of the two operations in \Cref{fig:ref_allocation_illustration}.
The algorithm continually applies one of the two operations until they can either no longer be applied or the budget is reached.
The algorithm will always terminate because \fadd\, increases the cost and utility and \fmove\, increases the utility by at least $\min_{l \in L} \fcost(l)$ and $\min_{l_1, l_2 \in L} |\futility(l_2)-\futility(l_1)|$, respectively.
Therefore either the budget will be filled or every segment will receive a reference from all vendors.

\begin{figure}[t]
\centering
\includegraphics[height=3.85cm]{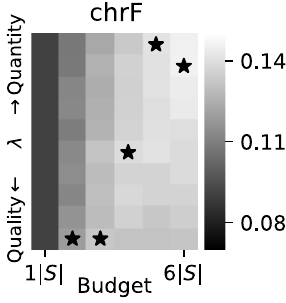}
\hfill
\includegraphics[height=3.85cm]{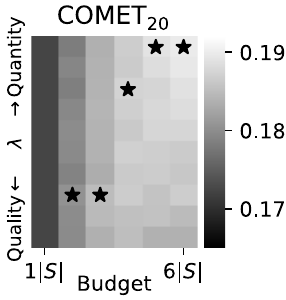}
\caption{Heatmaps of chrF (left) and \comet{20} (right) Kendall's $\tau$ correlations on reference configurations created with a specific budget (x-axis) and quality-quantity trade-off $\lambda$ (y-axis). $\bigstar$~marks the best value in each column (fixed budget).
The first column corresponds to the cheapest translation for all test segments, with no room for selection.
$\lambda \in [0, 0.7]$ and $t = 0.5$.
\observation
With a limited budget, e.g. $2|S|$ or $3|S|$, it makes more sense to add \emph{some} references of a higher quality rather than covering the whole test set with a second reference.
With more budget available, multiple references per segment become more beneficial.
}
\label{fig:budget_allocation}
\end{figure}

\begin{figure*}
\begin{minipage}{0.48\linewidth}
\includegraphics[width=\linewidth]{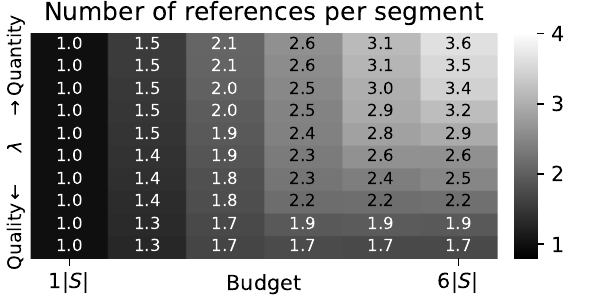}
\includegraphics[width=\linewidth]{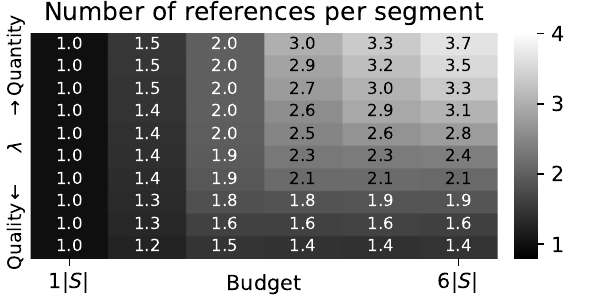}
\captionof{figure}{Average number of references per one segment allocated by \Cref{alg:ref_allocation} with $\tau=0.5$ (top) and $\tau=10^{-3}$ (bottom).}
\label{fig:budget_avgref}
\end{minipage}
\hfill
\begin{minipage}{0.48\linewidth}
\resizebox{\linewidth}{!}{
\begin{tabular}{clcccc}
\toprule
& \bf Metric & \bf R1\PE-R1 & \bf R2\PE-R2 & \bf R3\PE-R3 & \bf R4\PE-R4 \\
\midrule
\parbox[t]{2mm}{\multirow{6}{*}{\rotatebox[origin=c]{90}{\bf Layman PE}}} 
& BLEU         & \plusA 0.019 & \plusA 0.007 & \plusA 0.009 & \plusA 0.010 \\
& chrF         & \plusA 0.027 & \plusA 0.015 & \plusA 0.016 & \plusA 0.019 \\
& TER          & \plusA 0.026 & \plusA 0.015 & \plusA 0.013 & \plusA 0.014 \\
& COMET$^{20}$ & \plusA 0.016 & \plusA 0.011 & \plusA 0.010 & \plusA 0.009 \\
& COMET$^{22}$ & \plusA 0.008 & \plusA 0.006 & \plusA 0.007 & \plusA 0.005 \\
& BLEURT       & \plusA 0.017 & \plusA 0.015 & \plusA 0.011 & \plusA 0.010 \\
\midrule
\parbox[t]{2mm}{\multirow{6}{*}{\rotatebox[origin=c]{90}{\bf Student PE}}}
& BLEU         & \plusA 0.010 & \plusA 0.001 & \minusA 0.004 & \minusA 0.001 \\
& chrF         & \plusA 0.011 & \plusA 0.001 & \plusA 0.000 & \minusA 0.004 \\
& TER          & \plusA 0.003 & \plusA 0.001 & \plusA 0.005 & \minusA 0.002 \\
& COMET$^{20}$ & \plusA 0.010 & \plusA 0.003 & \minusA 0.002 & \minusA 0.002 \\
& COMET$^{22}$ & \plusA 0.002 & \plusA 0.001 & \plusA 0.000 & \minusA 0.002 \\
& BLEURT       & \plusA 0.021 & \plusA 0.022 & \plusA 0.005 & \minusA 0.004 \\
\midrule
\parbox[t]{2mm}{\multirow{6}{*}{\rotatebox[origin=c]{90}{\bf Prof. PE}}}
& BLEU         & \plusA 0.035 & \plusA 0.011 & \plusA 0.007 & \minusA 0.000 \\
& chrF         & \plusA 0.040 & \plusA 0.010 & \plusA 0.008 & \plusA 0.004 \\
& TER          & \plusA 0.023 & \plusA 0.014 & \plusA 0.003 & \minusA 0.002 \\
& COMET$^{20}$ & \plusA 0.016 & \plusA 0.006 & \plusA 0.000 & \plusA 0.005 \\
& COMET$^{22}$ & \plusA 0.008 & \plusA 0.002 & \plusA 0.003 & \plusA 0.005 \\
& BLEURT       & \plusA 0.027 & \plusA 0.022 & \plusA 0.005 & \minusA 0.001 \\
\bottomrule
\end{tabular}
}
\captionof{table}{Difference between using original translations (in \Cref{tab:ref_quality_orig}) and post-edited translations as references. Sections are divided based on who did the post-editing (layman, translatology student, or professional translator). This table expands on \Cref{tab:ref_quality_pe}. Absolute scores of individual reference subsets are in \Cref{tab:ref_quality_pe_fullplus}.}
\label{tab:ref_quality_pe_full}
\end{minipage}
\end{figure*}

In \Cref{fig:budget_allocation}, we show chrF and \comet{20} correlations when using the references selected by our algorithm.
The optimal preference between quality and quantity changes with increasing budget.
Using all of the budget on either quality or quantity would correspond to the bottom or top row, which are not optimal.
The best reference configurations for a particular budget, such as $|S|{\times}4$, four-times the price of the cheapest translation, contain a mixture of references from R1, R2, R3, and R4 with multiple references for some segments.
In addition to the metric correlations in \Cref{fig:budget_allocation}, we show the average number of references per source segment in \Cref{fig:budget_avgref}.
With focus on quality, each segment has fewer references.

\section{Qualitative Analysis}
\label{sec:qualitative_analysis}

In \Cref{tab:example_analysis,tab:example_analysis_2}, we show a single source segment, one system translation and multiple references and the metric scores.
BLEU ranges from 0 to 100 and both extremes are almost achieved just with a different reference.
The best human translation led to the lowest BLEU score because of a translation shift.
This is not surprising because BLEU operates on the surface-level.
Unexpectedly, a similar thing happens also with \comet{20}, which uses a distributed semantic representation of the segments.
This shows that parametric model-based metrics are not robust to changes in references.
In \Cref{tab:example_analysis_2}, the \comet{20} difference between references is large due to some translators deciding to drop the verb \textit{``spolupracovat'' (collaborate)}, which changes the meaning and the system translation is penalized.

\vspace{-2mm}
\section{Conclusion}

We showed that the quality of references is important for accurate automatic machine translation metrics.
The relationship is not straightforward: translatologists' translations, despite being the peak translation quality, are not the best references.
Rather, it is the \textit{standard commercial professional translations} that work best for current metrics.
The trend applies to both string-matching metrics as well as to parametric model-based ones.
Taking the \textit{average over multiple references provides the biggest benefit}, with diminishing returns after 7 references.
We also provided a heuristic-based \textit{algorithm for finding a good configuration of references given a budget}, which surpasses optimizing solely for quantity or quantity.

\vspace{-4mm}
\paragraph{Future work.}
The dataset size prevents system-level investigations.
Because there is little point in evaluating segments that are easy to translate, a follow-up approach could prioritize difficult-to translate segments.
This is used by \citet{isabelle-etal-2017-challenge} for creating a challenge set.
Future works should \textit{quantify} the references quality and ask how many segments are needed to fulfill a certain desideratum, such as effect size or metric accuracy.

\vspace{-4mm}
\paragraph{Limitations.}
We note the limitation of using a small dataset and a single language translation direction due to the costs of creating multiple rounds of high-quality references.
We are convinced the results hold in other scenarios as the effect directions are the same across multiple metrics and setups.

\clearpage

\begin{table*}[htbp]
\centering

\begin{minipage}[c]{0.29\textwidth}
\caption{BLEU and \comet{20} scores of the source ``\textit{\small Three Scottish students named among Europe's best}'' and the system translation ``\textit{\small Tři skotští studenti byli zařazeni mezi nejlepší v Evropě}''. Both metrics are multiplied by 100. \observation~All references are good translations but the scores vary.}
\label{tab:example_analysis}
\end{minipage}
\begin{minipage}[c]{0.70\textwidth}
\centering
\resizebox{0.9\linewidth}{!}{
\begin{tabular}{c<{\hspace{-3mm}}c<{\hspace{-4mm}}l}
\toprule
\bf BLEU & \bf \comet{20} & \,\,\,\bf Reference \\
\midrule
10 & 78 & K evropské špičce nově patří i tři skotští studenti \\
23 & 120 & Tři skotští studenti se umístili mezi nejlepšími v Evropě \\
23 & 121 & Tři skotští studenti mezi nejlepšími v Evropě \\
28 & 116 & Tři skotští studenti byli oceněni jako jedni z nejlepších v Evropě \\
28 & 115 & Tři skotští studenti byli jmenováni jako jedni z nejlepších v Evropě \\
28 & 114 & Tři skotští studenti byli vyhlášeni jako jedni z nejlepších v Evropě \\
32 & 117 & Tři skotští studenti byli jmenováni jednimi z nejlepších v Evropě \\
37 & 122 & Tři skotští studenti byli jmenováni mezi nejlepšími v Evropě \\
43 & 125 & Tři skotští studenti se zařadili mezi nejlepší v Evropě \\
43 & 121 & Tři skotští studenti patří mezi nejlepší v Evropě. \\
43 & 122 & Tři skotští studenti patří mezi nejlepší v Evropě \\
60 & 127 & Tři skotští studenti zařazeni mezi nejlepší v Evropě \\
100 & 131 & Tři skotští studenti byli zařazeni mezi nejlepší v Evropě \\
\bottomrule
\end{tabular}
}
\end{minipage}
\end{table*}

\begin{table*}[htbp]
\centering

\begin{minipage}[c]{0.29\textwidth}
\caption{BLEU and \comet{20} scores of the source ``\textit{\small Sony, Disney Back To Work On Third Spider-Man Film}'' and the system translation ``\textit{\small Disney se vrací, bude spolupracovat se Sony na třetím sólovém Spider-Man filmu}''. Both metrics are multiplied by 100. \observation~Some references omit part of the information and \comet{20} thus penalizes the system translation.}
\label{tab:example_analysis_2}
\end{minipage}
\begin{minipage}[c]{0.70\textwidth}
\resizebox{\linewidth}{!}{
\begin{tabular}{c<{\hspace{-3mm}}c<{\hspace{-4mm}}l}
\toprule
\bf BLEU & \bf \comet{20} & \,\,\,\bf Reference \\
\midrule
4 & -42 & Sony a Disney točí třetí film o Spidermanovi \\
4 & -33 & Sony a Disney točí třetí film o Spider-Manovi \\
8 & -9 & Sony a Disney pracují na třetím filmu o Spider-Manovi \\
8 & -4 & Sony a Disney pokračují v práci na třetím filmu o Spider-Manovi \\
8 & 1 & Sony a Disney opět pracují na třetím filmu o Spider-Manovi \\
8 & 15 & Sony a Disney spolupracují na třetím filmu o Spider-Manovi \\
4 & 16 & Sony a Disney budou spolupracovat při natáčení třetího filmo o Spider-manovi \\
8 & 28 & Sony a Disney opět spolupracují na třetím filmu o Spider-Manovi \\
8 & 30 & Sony a Disney budou spolupracovat na třetím filmu o Spider-Manovi \\
8 & 35 & Sony a Disney budou opět spolupracovat na třetím filmu o Spider-Manovi \\
17 & 52 & Disney bude znovu spolupracovat se společností Sony na třetím filmu Spider-Man \\
10 & 64 & Disney bude se Sony dál pracovat na třetím filmu se Spider-Manem \\
50 & 73 & Disney bude spolupracovat se Sony na třetím sólovém filmu o Spider-Manovi \\
75 & 99 & Disney se vrací, bude spolupracovat se Sony na třetím filmu o Spider-Manovi \\
78 & 106 & Disney se vrací, bude spolupracovat se Sony na třetím sólovém filu Spider-Man. \\
79 & 108 & Disney se vrací, bude spolupracovat se Sony na třetím Spider-Man filmu \\
100 & 121 & Disney se vrací, bude spolupracovat se Sony na třetím sólovém Spider-Man filmu \\
\bottomrule
\end{tabular}
}
\end{minipage}
\vspace{-1mm}
\end{table*}

\begin{table*}[htbp]
\centering
\resizebox{\linewidth}{!}{
\begin{tabular}{clccccccccccccccc}
\toprule
&
& \bf R1
& \bf R2
& \bf R3
& \bf R4
& \bf R\{1,2\}
& \bf R\{1,3\}
& \bf R\{1,4\}
& \bf R\{2,3\}
& \bf R\{2,4\}
& \bf R\{3,4\}
& \bf R\{1,2,3\}
& \bf R\{1,2,4\}
& \bf R\{1,3,4\}
& \bf R\{2,3,4\}
& \bf Rx
\\
\midrule
\parbox[t]{2mm}{\multirow{7}{*}{\rotatebox[origin=c]{90}{\bf Average}}}
& BLEU         & 0.082 \blackindicator{0.09} & 0.103 \blackindicator{0.24} & 0.109 \blackindicator{0.28} & 0.103 \blackindicator{0.24} & 0.109 \blackindicator{0.28} & 0.122 \blackindicator{0.37} & 0.114 \blackindicator{0.31} & 0.132 \blackindicator{0.44} & 0.124 \blackindicator{0.39} & 0.114 \blackindicator{0.32} & 0.136 \blackindicator{0.47} & 0.124 \blackindicator{0.39} & 0.125 \blackindicator{0.39} & 0.130 \blackindicator{0.43} & 0.134 \blackindicator{0.46} \\
& chrF         & 0.090 \blackindicator{0.15} & 0.125 \blackindicator{0.39} & 0.128 \blackindicator{0.41} & 0.123 \blackindicator{0.38} & 0.121 \blackindicator{0.36} & 0.135 \blackindicator{0.47} & 0.124 \blackindicator{0.38} & 0.148 \blackindicator{0.56} & 0.139 \blackindicator{0.49} & 0.135 \blackindicator{0.47} & 0.146 \blackindicator{0.54} & 0.135 \blackindicator{0.46} & 0.140 \blackindicator{0.50} & 0.147 \blackindicator{0.55} & 0.147 \blackindicator{0.55} \\
& TER          & 0.082 \blackindicator{0.08} & 0.092 \blackindicator{0.16} & 0.114 \blackindicator{0.31} & 0.105 \blackindicator{0.25} & 0.095 \blackindicator{0.18} & 0.120 \blackindicator{0.36} & 0.107 \blackindicator{0.26} & 0.125 \blackindicator{0.40} & 0.117 \blackindicator{0.33} & 0.120 \blackindicator{0.35} & 0.121 \blackindicator{0.37} & 0.110 \blackindicator{0.29} & 0.123 \blackindicator{0.38} & 0.127 \blackindicator{0.40} & 0.124 \blackindicator{0.39} \\
& \comet{20} & 0.172 \blackindicator{0.73} & 0.176 \blackindicator{0.76} & 0.185 \blackindicator{0.82} & 0.181 \blackindicator{0.79} & 0.181 \blackindicator{0.79} & 0.189 \blackindicator{0.85} & 0.185 \blackindicator{0.82} & 0.191 \blackindicator{0.86} & 0.183 \blackindicator{0.81} & 0.188 \blackindicator{0.84} & 0.190 \blackindicator{0.85} & 0.185 \blackindicator{0.82} & 0.190 \blackindicator{0.85} & 0.189 \blackindicator{0.85} & 0.189 \blackindicator{0.85} \\
& \comet{22} & 0.189 \blackindicator{0.85} & 0.195 \blackindicator{0.90} & 0.191 \blackindicator{0.86} & 0.192 \blackindicator{0.87} & 0.195 \blackindicator{0.90} & 0.197 \blackindicator{0.91} & 0.194 \blackindicator{0.89} & 0.201 \blackindicator{0.93} & 0.197 \blackindicator{0.91} & 0.195 \blackindicator{0.89} & 0.200 \blackindicator{0.93} & 0.197 \blackindicator{0.91} & 0.197 \blackindicator{0.91} & 0.199 \blackindicator{0.92} & 0.199 \blackindicator{0.92} \\
& BLEURT       & 0.159 \blackindicator{0.64} & 0.156 \blackindicator{0.61} & 0.199 \blackindicator{0.92} & 0.178 \blackindicator{0.77} & 0.171 \blackindicator{0.72} & 0.203 \blackindicator{0.95} & 0.183 \blackindicator{0.81} & 0.201 \blackindicator{0.94} & 0.180 \blackindicator{0.79} & 0.203 \blackindicator{0.95} & 0.201 \blackindicator{0.94} & 0.184 \blackindicator{0.81} & 0.204 \blackindicator{0.96} & 0.202 \blackindicator{0.94} & 0.202 \blackindicator{0.94} \\
\cmidrule{2-17}
& \bf Average & 0.129 \blackindicator{0.42} & 0.141 \blackindicator{0.51} & 0.154 \blackindicator{0.60} & 0.147 \blackindicator{0.55} & 0.145 \blackindicator{0.54} & 0.161 \blackindicator{0.65} & 0.151 \blackindicator{0.58} & 0.166 \blackindicator{0.69} & 0.157 \blackindicator{0.62} & 0.159 \blackindicator{0.64} & 0.166 \blackindicator{0.68} & 0.156 \blackindicator{0.61} & 0.163 \blackindicator{0.66} & 0.166 \blackindicator{0.68} & 0.166 \blackindicator{0.68} \\
\midrule
\parbox[t]{2mm}{\multirow{7}{*}{\rotatebox[origin=c]{90}{\bf Max}}}
& BLEU         & 0.082 \blackindicator{0.09} & 0.103 \blackindicator{0.24} & 0.109 \blackindicator{0.28} & 0.103 \blackindicator{0.24} & 0.116 \blackindicator{0.33} & 0.122 \blackindicator{0.37} & 0.118 \blackindicator{0.34} & 0.135 \blackindicator{0.46} & 0.132 \blackindicator{0.44} & 0.111 \blackindicator{0.29} & 0.138 \blackindicator{0.48} & 0.137 \blackindicator{0.48} & 0.121 \blackindicator{0.36} & 0.135 \blackindicator{0.46} & 0.137 \blackindicator{0.48} \\
& chrF         & 0.090 \blackindicator{0.15} & 0.125 \blackindicator{0.39} & 0.128 \blackindicator{0.41} & 0.123 \blackindicator{0.38} & 0.133 \blackindicator{0.45} & 0.137 \blackindicator{0.48} & 0.129 \blackindicator{0.42} & 0.139 \blackindicator{0.50} & 0.146 \blackindicator{0.54} & 0.135 \blackindicator{0.46} & 0.144 \blackindicator{0.53} & 0.148 \blackindicator{0.56} & 0.140 \blackindicator{0.50} & 0.144 \blackindicator{0.53} & 0.147 \blackindicator{0.55} \\
& TER          & 0.082 \blackindicator{0.08} & 0.092 \blackindicator{0.16} & 0.114 \blackindicator{0.31} & 0.105 \blackindicator{0.25} & 0.101 \blackindicator{0.22} & 0.124 \blackindicator{0.39} & 0.116 \blackindicator{0.33} & 0.132 \blackindicator{0.44} & 0.128 \blackindicator{0.42} & 0.117 \blackindicator{0.34} & 0.132 \blackindicator{0.44} & 0.130 \blackindicator{0.43} & 0.126 \blackindicator{0.40} & 0.135 \blackindicator{0.46} & 0.134 \blackindicator{0.46} \\
& COMET$^{20}$ & 0.172 \blackindicator{0.73} & 0.176 \blackindicator{0.76} & 0.185 \blackindicator{0.82} & 0.181 \blackindicator{0.79} & 0.177 \blackindicator{0.76} & 0.184 \blackindicator{0.81} & 0.185 \blackindicator{0.82} & 0.180 \blackindicator{0.79} & 0.183 \blackindicator{0.81} & 0.183 \blackindicator{0.81} & 0.181 \blackindicator{0.79} & 0.183 \blackindicator{0.81} & 0.184 \blackindicator{0.81} & 0.181 \blackindicator{0.79} & 0.182 \blackindicator{0.80} \\
& COMET$^{22}$ & 0.189 \blackindicator{0.85} & 0.195 \blackindicator{0.90} & 0.191 \blackindicator{0.86} & 0.192 \blackindicator{0.87} & 0.195 \blackindicator{0.89} & 0.191 \blackindicator{0.86} & 0.196 \blackindicator{0.90} & 0.191 \blackindicator{0.87} & 0.195 \blackindicator{0.90} & 0.191 \blackindicator{0.87} & 0.192 \blackindicator{0.87} & 0.197 \blackindicator{0.90} & 0.191 \blackindicator{0.87} & 0.192 \blackindicator{0.87} & 0.192 \blackindicator{0.87} \\
& BLEURT       & 0.159 \blackindicator{0.64} & 0.156 \blackindicator{0.61} & 0.199 \blackindicator{0.92} & 0.178 \blackindicator{0.77} & 0.180 \blackindicator{0.79} & 0.199 \blackindicator{0.92} & 0.190 \blackindicator{0.86} & 0.188 \blackindicator{0.85} & 0.181 \blackindicator{0.80} & 0.193 \blackindicator{0.88} & 0.197 \blackindicator{0.91} & 0.192 \blackindicator{0.87} & 0.200 \blackindicator{0.93} & 0.188 \blackindicator{0.84} & 0.197 \blackindicator{0.91} \\
\cmidrule{2-17}
& \bf Average & 0.129 \blackindicator{0.42} & 0.141 \blackindicator{0.51} & 0.154 \blackindicator{0.60} & 0.147 \blackindicator{0.55} & 0.150 \blackindicator{0.57} & 0.159 \blackindicator{0.64} & 0.156 \blackindicator{0.61} & 0.161 \blackindicator{0.65} & 0.161 \blackindicator{0.65} & 0.155 \blackindicator{0.61} & 0.164 \blackindicator{0.67} & 0.164 \blackindicator{0.67} & 0.160 \blackindicator{0.64} & 0.162 \blackindicator{0.66} & 0.165 \blackindicator{0.68} \\
\bottomrule
\end{tabular}
}
\caption{Comparison using either a single or multiple references and taking the average or maximum on segment-level. This table expands on \Cref{tab:ref_quantity}.
The black boxes indicate the reported value of Kendall's $\tau$ visually and are comparable across columns as well as rows.
}
\label{tab:ref_quantity_full}
\end{table*}

\begin{table*}[htbp]
\centering
\resizebox{0.9\linewidth}{!}{
\begin{tabular}{clcccccccccc}
\toprule
&
& \bf R1\PE
& \bf R2\PE
& \bf R3\PE
& \bf R4\PE
& \bf Rx\PE
& \bf R\{1,1\PEx\}
& \bf R\{2,2\PEx\}
& \bf R\{3,3\PEx\}
& \bf R\{4,4\PEx\}
& \bf R\{x,x\PEx\}
\\
\midrule
\parbox[t]{2mm}{\multirow{7}{*}{\rotatebox[origin=c]{90}{\bf Layman PE}}}
& BLEU         & 0.101 \blackindicator{0.22} & 0.111 \blackindicator{0.29} & 0.117 \blackindicator{0.34} & 0.113 \blackindicator{0.31} & 0.144 \blackindicator{0.53} & 0.092 \blackindicator{0.16} & 0.107 \blackindicator{0.27} & 0.113 \blackindicator{0.31} & 0.108 \blackindicator{0.27} & 0.140 \blackindicator{0.50} \\
& chrF         & 0.118 \blackindicator{0.34} & 0.139 \blackindicator{0.49} & 0.144 \blackindicator{0.53} & 0.142 \blackindicator{0.51} & 0.164 \blackindicator{0.67} & 0.106 \blackindicator{0.26} & 0.134 \blackindicator{0.45} & 0.137 \blackindicator{0.48} & 0.135 \blackindicator{0.47} & 0.159 \blackindicator{0.64} \\
& TER          & 0.107 \blackindicator{0.27} & 0.107 \blackindicator{0.27} & 0.127 \blackindicator{0.41} & 0.119 \blackindicator{0.35} & 0.142 \blackindicator{0.52} & 0.099 \blackindicator{0.21} & 0.102 \blackindicator{0.23} & 0.123 \blackindicator{0.38} & 0.116 \blackindicator{0.33} & 0.139 \blackindicator{0.49} \\
& COMET$^{20}$ & 0.188 \blackindicator{0.84} & 0.187 \blackindicator{0.83} & 0.195 \blackindicator{0.89} & 0.190 \blackindicator{0.85} & 0.198 \blackindicator{0.91} & 0.183 \blackindicator{0.81} & 0.184 \blackindicator{0.81} & 0.193 \blackindicator{0.88} & 0.188 \blackindicator{0.84} & 0.197 \blackindicator{0.90} \\
& COMET$^{22}$ & 0.197 \blackindicator{0.91} & 0.201 \blackindicator{0.94} & 0.198 \blackindicator{0.91} & 0.197 \blackindicator{0.91} & 0.203 \blackindicator{0.95} & 0.195 \blackindicator{0.89} & 0.200 \blackindicator{0.93} & 0.196 \blackindicator{0.90} & 0.196 \blackindicator{0.90} & 0.202 \blackindicator{0.95} \\
& BLEURT       & 0.176 \blackindicator{0.76} & 0.170 \blackindicator{0.72} & 0.210 \blackindicator{1.00} & 0.188 \blackindicator{0.84} & 0.209 \blackindicator{1.00} & 0.169 \blackindicator{0.71} & 0.165 \blackindicator{0.68} & 0.206 \blackindicator{0.97} & 0.186 \blackindicator{0.83} & 0.209 \blackindicator{0.99} \\
\cmidrule{2-12}
& \bf Average & 0.148 \blackindicator{0.56} & 0.153 \blackindicator{0.59} & 0.165 \blackindicator{0.68} & 0.158 \blackindicator{0.63} & 0.177 \blackindicator{0.76} & 0.141 \blackindicator{0.51} & 0.149 \blackindicator{0.56} & 0.161 \blackindicator{0.65} & 0.155 \blackindicator{0.61} & 0.174 \blackindicator{0.75} \\
\midrule
\parbox[t]{2mm}{\multirow{7}{*}{\rotatebox[origin=c]{90}{\bf Student PE}}} 
& BLEU         & 0.092 \blackindicator{0.16} & 0.104 \blackindicator{0.24} & 0.105 \blackindicator{0.25} & 0.102 \blackindicator{0.23} & 0.123 \blackindicator{0.38} & 0.089 \blackindicator{0.14} & 0.107 \blackindicator{0.26} & 0.108 \blackindicator{0.27} & 0.103 \blackindicator{0.24} & 0.130 \blackindicator{0.43} \\
& chrF         & 0.102 \blackindicator{0.23} & 0.126 \blackindicator{0.40} & 0.128 \blackindicator{0.41} & 0.119 \blackindicator{0.35} & 0.139 \blackindicator{0.49} & 0.097 \blackindicator{0.20} & 0.127 \blackindicator{0.41} & 0.130 \blackindicator{0.43} & 0.122 \blackindicator{0.37} & 0.144 \blackindicator{0.53} \\
& TER          & 0.085 \blackindicator{0.10} & 0.093 \blackindicator{0.16} & 0.119 \blackindicator{0.35} & 0.103 \blackindicator{0.23} & 0.119 \blackindicator{0.35} & 0.084 \blackindicator{0.10} & 0.095 \blackindicator{0.18} & 0.119 \blackindicator{0.35} & 0.104 \blackindicator{0.24} & 0.123 \blackindicator{0.38} \\
& COMET$^{20}$ & 0.182 \blackindicator{0.80} & 0.179 \blackindicator{0.78} & 0.183 \blackindicator{0.81} & 0.179 \blackindicator{0.78} & 0.186 \blackindicator{0.83} & 0.179 \blackindicator{0.78} & 0.179 \blackindicator{0.78} & 0.186 \blackindicator{0.83} & 0.181 \blackindicator{0.79} & 0.188 \blackindicator{0.85} \\
& COMET$^{22}$ & 0.191 \blackindicator{0.87} & 0.196 \blackindicator{0.90} & 0.191 \blackindicator{0.86} & 0.189 \blackindicator{0.85} & 0.195 \blackindicator{0.89} & 0.191 \blackindicator{0.87} & 0.197 \blackindicator{0.91} & 0.193 \blackindicator{0.88} & 0.191 \blackindicator{0.86} & 0.197 \blackindicator{0.91} \\
& BLEURT       & 0.180 \blackindicator{0.79} & 0.178 \blackindicator{0.77} & 0.203 \blackindicator{0.95} & 0.174 \blackindicator{0.74} & 0.199 \blackindicator{0.92} & 0.172 \blackindicator{0.73} & 0.170 \blackindicator{0.71} & 0.204 \blackindicator{0.96} & 0.177 \blackindicator{0.76} & 0.202 \blackindicator{0.94} \\
\cmidrule{2-12}
& \bf Average & 0.139 \blackindicator{0.49} & 0.146 \blackindicator{0.54} & 0.155 \blackindicator{0.61} & 0.144 \blackindicator{0.53} & 0.160 \blackindicator{0.64} & 0.136 \blackindicator{0.47} & 0.146 \blackindicator{0.54} & 0.157 \blackindicator{0.62} & 0.146 \blackindicator{0.54} & 0.164 \blackindicator{0.67} \\
\midrule
\parbox[t]{2mm}{\multirow{7}{*}{\rotatebox[origin=c]{90}{\bf Professional PE}}} 
& BLEU         & 0.118 \blackindicator{0.34} & 0.114 \blackindicator{0.31} & 0.115 \blackindicator{0.32} & 0.103 \blackindicator{0.24} & 0.127 \blackindicator{0.41} & 0.103 \blackindicator{0.24} & 0.113 \blackindicator{0.30} & 0.113 \blackindicator{0.31} & 0.104 \blackindicator{0.24} & 0.133 \blackindicator{0.45} \\
& chrF         & 0.131 \blackindicator{0.43} & 0.135 \blackindicator{0.46} & 0.136 \blackindicator{0.47} & 0.127 \blackindicator{0.40} & 0.146 \blackindicator{0.54} & 0.113 \blackindicator{0.31} & 0.133 \blackindicator{0.45} & 0.135 \blackindicator{0.47} & 0.126 \blackindicator{0.40} & 0.149 \blackindicator{0.56} \\
& TER          & 0.105 \blackindicator{0.25} & 0.106 \blackindicator{0.25} & 0.116 \blackindicator{0.33} & 0.103 \blackindicator{0.24} & 0.118 \blackindicator{0.34} & 0.094 \blackindicator{0.17} & 0.102 \blackindicator{0.23} & 0.118 \blackindicator{0.34} & 0.104 \blackindicator{0.25} & 0.122 \blackindicator{0.37} \\
& COMET$^{20}$ & 0.188 \blackindicator{0.84} & 0.182 \blackindicator{0.80} & 0.185 \blackindicator{0.82} & 0.186 \blackindicator{0.83} & 0.190 \blackindicator{0.86} & 0.183 \blackindicator{0.81} & 0.181 \blackindicator{0.79} & 0.189 \blackindicator{0.85} & 0.185 \blackindicator{0.82} & 0.191 \blackindicator{0.87} \\
& COMET$^{22}$ & 0.198 \blackindicator{0.91} & 0.198 \blackindicator{0.91} & 0.195 \blackindicator{0.89} & 0.196 \blackindicator{0.90} & 0.199 \blackindicator{0.92} & 0.195 \blackindicator{0.89} & 0.199 \blackindicator{0.92} & 0.195 \blackindicator{0.89} & 0.195 \blackindicator{0.89} & 0.200 \blackindicator{0.93} \\
& BLEURT       & 0.186 \blackindicator{0.83} & 0.178 \blackindicator{0.77} & 0.204 \blackindicator{0.96} & 0.176 \blackindicator{0.76} & 0.197 \blackindicator{0.91} & 0.177 \blackindicator{0.76} & 0.172 \blackindicator{0.73} & 0.206 \blackindicator{0.97} & 0.179 \blackindicator{0.78} & 0.202 \blackindicator{0.94} \\
\cmidrule{2-12}
& \bf Average & 0.154 \blackindicator{0.60} & 0.152 \blackindicator{0.59} & 0.159 \blackindicator{0.63} & 0.149 \blackindicator{0.56} & 0.163 \blackindicator{0.66} & 0.144 \blackindicator{0.53} & 0.150 \blackindicator{0.57} & 0.159 \blackindicator{0.64} & 0.149 \blackindicator{0.56} & 0.166 \blackindicator{0.69} \\
\bottomrule
\end{tabular}
}
\caption{Metric performance when using post-edited references also jointly with their original versions (averaged at the segment-level). This table expands on \Cref{tab:ref_quality_pe,tab:ref_quality_pe_full}.}
\label{tab:ref_quality_pe_fullplus}
\end{table*}

\clearpage

\section*{Acknowledgements}

We extend our gratitude to Yasmin Moslem, Isabella Lai, Theia Vogel, Blanka Sokołowska, Raj Dabre, Tom Kocmi, and Farhan Samir, who proofread this paper in their free time.
Ondřej Bojar received funding from Ministry of Education, Youth and Sports of the Czech Republic LM2018101 LINDAT/CLARIAH-CZ and the 19-26934X grant of the Czech Science Foundation (NEUREM3).

\section*{Bibliographical References}
\vspace*{-5mm}

\bibliography{misc/bibliography}
\bibliographystyle{misc/acl_natbib}

\end{document}